\documentclass[runningheads]{llncs}

\usepackage[usenames,dvipsnames]{xcolor}
\usepackage{graphicx}

\usepackage{tikz}
\usepackage{comment}
\usepackage{amsmath,amssymb} 

\usepackage{wrapfig}
\usepackage{mathtools}
\usepackage{scalerel}
\usepackage{amssymb}
\usepackage{booktabs}
\usepackage{multirow}
\usepackage{xspace}
\usepackage{pifont}
\usepackage{hyperref}

\newcommand{\cmark}{\ding{51}}%
\newcommand{\xmark}{\ding{55}}%

\newcommand{\camera}[1]{{\textcolor{ForestGreen}{\textbf{#1}}\xspace}}
\newcommand{\structure}[1]{{\textcolor{RoyalBlue}{\textbf{#1}}\xspace}}
\newcommand{\scale}[1]{{\textcolor{orange}{\textbf{#1}}\xspace}}

\newcommand{\etal}{\textit{et al}.}
\newcommand{\ie}{\textit{i}.\textit{e}.}
\newcommand{\eg}{\textit{e}.\textit{g}.}
\newcommand{\nerfw}{\mbox{NeRF-W}\xspace}

\usepackage[accsupp]{axessibility}  

\begin{document}
\pagestyle{headings}
\mainmatter
\def\ECCVSubNumber{****}  

\title{The One Where They Reconstructed \\
3D Humans and Environments in TV Shows} %

\titlerunning{Reconstructing 3D Humans and Environments in TV Shows}
%
\author{
Georgios Pavlakos$^*$, 
Ethan Weber$^*$, 
Matthew Tancik, 
Angjoo Kanazawa
}
\authorrunning{G. Pavlakos$^*$, E. Weber$^*$, M. Tancik, A. Kanazawa}
%
\institute{University of California, Berkeley}

\maketitle
\begin{figure}[h]
\centering
\includegraphics[width=\textwidth]{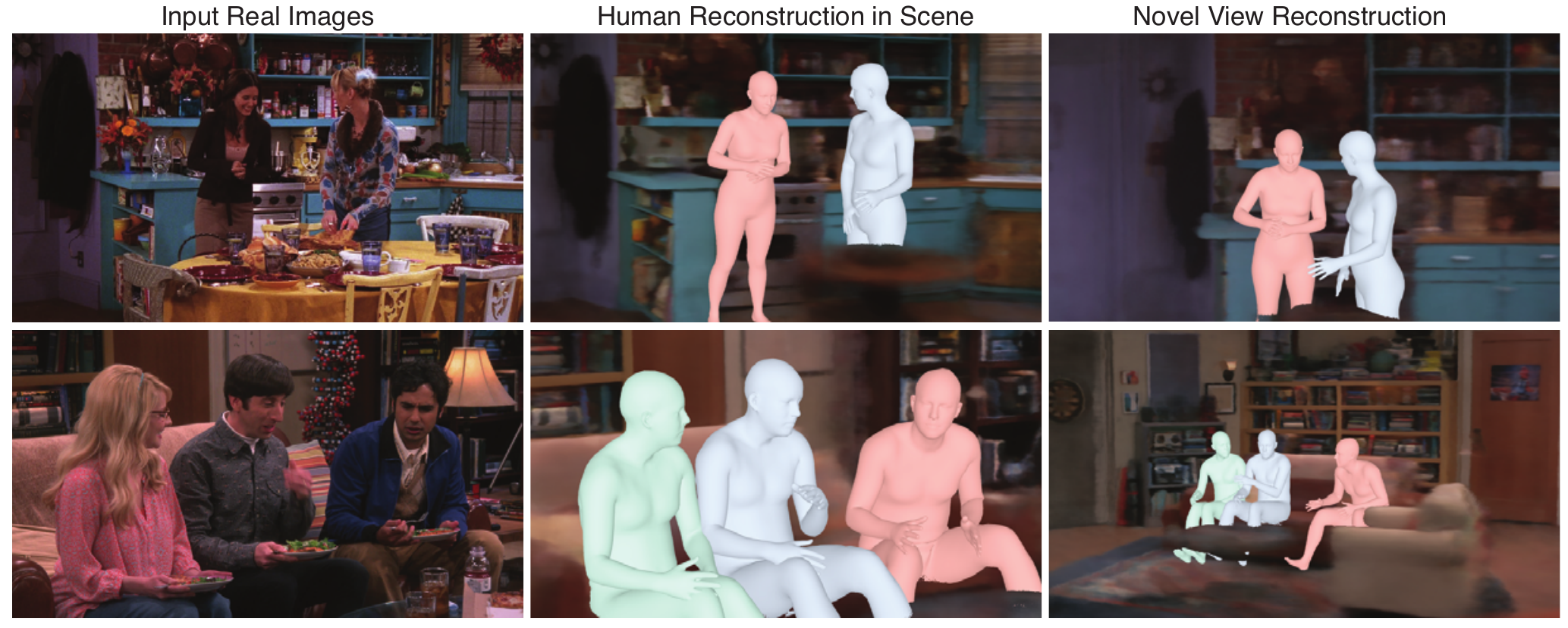}
\caption{\textbf{Reconstruction of humans in TV show environments.} Given images across the whole season of a TV show, we present an approach that recovers the 3D scene context, which enables accurate estimation of every actor's 3D pose and location. We show the input (left), the mesh reconstructions of the actors in the camera view (center) and in a novel view (right). Human meshes are visualized against the reconstructed scene, which is represented by a Neural Radiance Field (NeRF). To appreciate the correct 3D localization of people, notice the position in the novel view and the occlusions. Readers are encouraged to watch video results in the project page: \url{http://ethanweber.me/sitcoms3D/}.}
\label{fig:teaser}
\end{figure}

{\renewcommand{\thefootnote}{\fnsymbol{footnote}} \footnotetext[1]{Equal contribution.}}

\newcommand{\todo}[1]{\textcolor{red}{[#1]}}

\newcommand{\mypar}[1]{\vspace{1mm}\noindent{\textbf{#1}}}

\newcommand{\Csmall}[1]{{\scaleto{C}{3.7pt}}}
\newcommand{\CWsmall}[1]{{\scaleto{CW}{3.7pt}}}

\begin{abstract} 
TV shows depict a wide variety of human behaviors and have been studied extensively for their potential to be a rich source of data for many applications. However, the majority of the existing work focuses on 2D recognition tasks. In this paper, we make the observation that there is a certain persistence in TV shows, \ie, repetition of the environments and the humans, which makes possible the 3D reconstruction of this content. Building on this insight, we propose an automatic approach that operates on an entire season of a TV show and aggregates information in 3D; we build a 3D model of the environment, compute camera information, static 3D scene structure and body scale information. Then, we demonstrate how this information acts as {\it rich 3D context} that can guide and improve the recovery of 3D human pose and position in these environments. Moreover, we show that reasoning about humans and their environment in 3D enables a broad range of downstream applications: re-identification, gaze estimation, cinematography and image editing. We apply our approach on environments from seven iconic TV shows and perform an extensive evaluation of the proposed system.
\end{abstract}
\section{Introduction}
\label{sec:intro}
Remember that time when you binge-watched an entire season of your favorite TV show, \eg, ``Friends'', over a weekend? After that experience, you would know the layout of the rooms, the locations of the furniture, and even the relative height of the characters as they interact closely on screen. As a result, for any frame, you could tell where the room is viewed from, where the characters are situated, and how they relate to the rest of the scene, even the parts of the scene outside the frame. Essentially, as viewers, we aggregate all the visual information into a dynamic 3D world where the new observations are aligned to. 

In this paper, we propose a method that can similarly aggregate 3D information over video collections and use it to perceive accurate 3D human pose and location of the actors. Although reconstruction of dynamic scenes is challenging from a single video clip, our insight is that in the context of TV shows, across many episodes, there are many video clips that \textit{depict the same scene and people many times.} The repeated observations provide a strong multi-view signal of the underlying scene, enabling reconstruction of the camera and the dense structure. These serve as context to accurately recover the 3D pose and location of the people in the 3D environment. A representative result is shown in Figure~\ref{fig:teaser}. Although we demonstrate our method \& results on TV shows, our insight is also applicable to other domains with repetition in the environment and the people, \eg, sports~\cite{homayounfar2017sports,rematas2018soccer,zhu2020reconstructing}, late night shows~\cite{ginosar2019learning,ng2021body2hands} and movies~\cite{pavlakos2020human}.

\begin{figure*}[t]
  \centering
  \includegraphics[width=\linewidth]{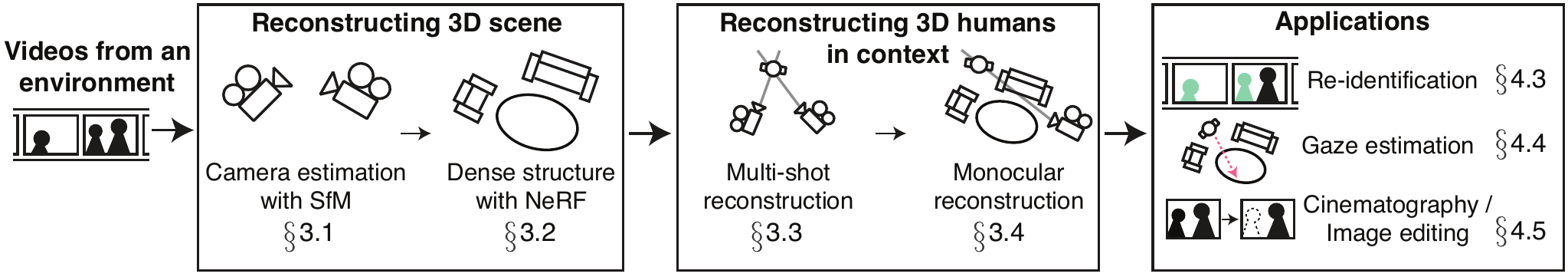}
  \caption{\textbf{Overview of our workflow.} First, we use a collection of videos from a TV show environment and reconstruct the 3D scene (cameras and dense structure). We then use this information to recover accurate 3D pose and location of people over shot boundaries and on monocular frames. The recovered 3D information is immediately useful for various downstream applications.}
  \label{fig:overview}
\end{figure*}

We operationalize our insight by focusing on an entire season from TV shows and collecting the sequences that correspond to a specific environment. These sequences are organized in shots~\cite{arijon1991grammar}, which are typically captured by different cameras. To collect a diverse set of images, we sample frames at the shot boundaries (cuts between cameras). This ensures a wide variety of viewpoints, while avoiding redundancy, making it practical to apply a structure-from-motion pipeline~\cite{schonberger2016structure} to estimate the intrinsic and extrinsic camera parameters (calibration). Then we use a neural radiance field (\nerfw~\cite{martin2021nerf}) to disentangle the static and transient components and obtain a dense 3D reconstruction (Figure~\ref{fig:overview}, first block).

The 3D scene reconstruction offers rich 3D context - cameras and scene structure - enabling an in-depth study of humans (Figure~\ref{fig:overview}, second block). First, for frames on the shot boundaries, the viewpoints from the two different shots act as effective multi-view (or multi-shot) information for human reconstruction~\cite{pavlakos2020human}. We use the calibrated cameras and propose a multi-shot human reconstruction method, which jointly solves for body pose, body shape, identity and location. In this {\bf calibrated multi-shot} method, camera information enables triangulation of people, which removes ambiguity and provides significant improvement upon the equivalent uncalibrated baseline~\cite{pavlakos2020human}. Next, human reconstructions on the shot boundary inform us of the scale of each person relative to the scene. This is additional 3D context that is complementary to the cameras and scene structure. Since most frames are not on shot boundaries, we also formulate a monocular human reconstruction method that is explicitly guided by the extracted 3D context (camera, structure, body scale). The successful integration of the 3D context in our {\bf contextual monocular} method leads to improvements over the state-of-the-art monocular baselines.

Our proposed ``3Dification'' of TV shows opens the door to many immediate applications (Figure~\ref{fig:overview}, third block). First, our human reconstruction on the shot boundaries associates person detections, by incorporating geometric and anthropometric constraints. We show that this form of {\bf re-identification} consistently outperforms traditional image-based baselines~\cite{fu2021unsupervised,huang2018person,huang2020movienet}. In parallel, from our reconstructed humans we can extract reliable {\bf gaze information}, which can outperform specialized gaze estimation~\cite{recasens2017following}. Moreover, our results provide estimates of the camera-to-person distance, which is relevant for {\bf cinematography} applications~\cite{savardi2021cinescale,savardi2018shot}. Finally, we illustrate the potential use in {\bf image editing} applications, like object insertion or human deletion.

In summary, our contributions can be summarized as follows:
\begin{itemize}
    \item We identify the significant amount of 3D context (cameras, structure and body shape) in domains with repetition in the environment and the people, \eg, TV shows, and propose a method to aggregate it from video sequences.
    \item We propose a formulation that integrates this context in 3D human estimation methods, which improves human reconstruction.
    \item We demonstrate how the aggregated 3D information can help a wide variety of downstream tasks: re-ID, gaze estimation, cinematography, image editing.
    \item We perform extensive qualitative and quantitative evaluation to validate the quality of our recovered 3D results.
\end{itemize}
\section{Related work}
\label{sec:related}

\subsection{Perceiving TV shows}
The computer vision community has a long history of works on perceiving TV shows/movies. One of the most common tasks in this setting is studying the show characters with emphasis in face/character identification~\cite{arandjelovic2005automatic,everingham2005identifying,nagrani2018benedict,parkhi2018automated,sivic2009you,tapaswi2019video}, where different cues have also been explored, \eg, body, voice or gaze~\cite{brown2021face,everingham2006hello,marin2021laeopp,marin2014detecting}. TV show data has been used extensively to study human behavior. Ferrari~\etal~study 2D pose estimation~\cite{ferrari2008progressive} and perform pose-based analysis~\cite{ferrari2009pose}. Patron~\etal~\cite{patron2012structured} and Hoai~\etal~\cite{hoai2014talking} focus on human interactions, while Recasens~\etal~\cite{recasens2017following} and Mar{\'\i}n-Jim{\'e}nez~\etal~\cite{marin2021laeopp} use this data to study gaze. Vondrick~\etal~\cite{vondrick2016anticipating} use sequences from TV shows to learn activity forecasting, while Wang~\etal~\cite{wang2017binge} leverage it for affordance learning. Despite this attention, all the above methods reason in 2D, with only a few exceptions. Everingham and Zisserman~\cite{everingham2005identifying} use a 3D head model for re-identification. Here, we demonstrate how 3D location information can significantly simplify the re-ID problem. Pavlakos~\etal~\cite{pavlakos2020human} reconstruct humans from videos with multiple shots. However, they operate without camera calibration, while we show the importance of recovering reliable cameras.

\begin{wrapfigure}{R}{0.5\textwidth}
  \centering
  \includegraphics[width=\linewidth]{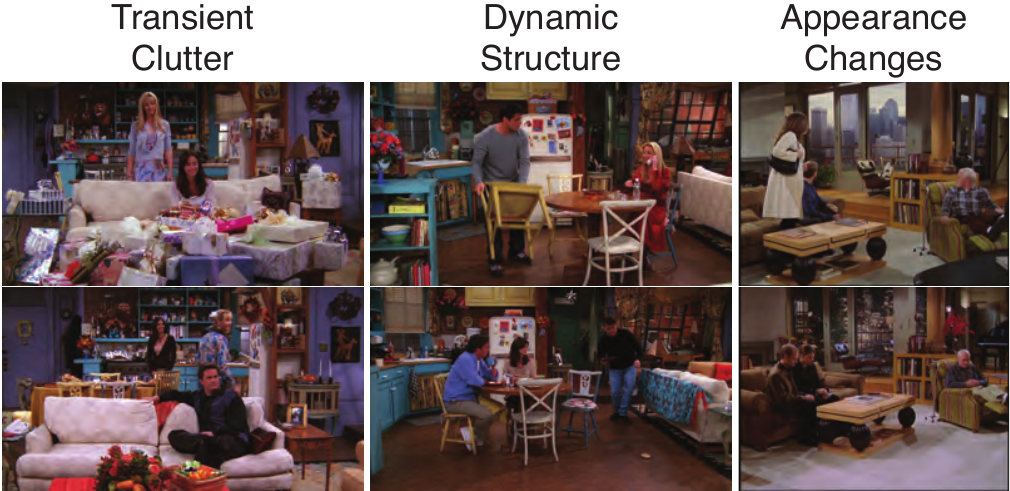}
  \caption{\textbf{Reconstruction challenges} of TV shows include transient and dynamic objects as well as appearance changes. 
  }
  \label{fig:in_the_wild}
\end{wrapfigure}

\subsection{Scene reconstruction}
Reconstruction of 3D scenes is a well studied problem, \eg,~\cite{agarwal2011building,huang2018deepmvs,schonberger2016structure,schonberger2016pixelwise}, however, most methods assume static scenes. Related work focuses on dynamic reconstruction~\cite{ballan2010unstructured,mustafa2021temporally}, but requires capture from multiple wide-baseline synchronized cameras. Luo~\etal~\cite{luo2020consistent} and Kopf~\etal~\cite{kopf2021robust} present pipelines for recovering depth in monocular videos that include humans, but they assume that the underlying scene is static. View synthesis approaches like NeRF~\cite{mildenhall2020nerf} and follow-ups~\cite{oechsle2021unisurf,yariv2021volume} can be used to solve multi-view stereo, however these also assume static scenes. Other extensions of NeRF focus on reconstructing 3D motion in the scene~\cite{gao2021dynamic,li2021neural,park2021nerfies}, but are often limited when handling changes in appearances and transient objects. NeRF in the wild~\cite{martin2021nerf} is the most relevant approach for the type of data we use (Figure~\ref{fig:in_the_wild}), since it can deal with appearance and transient changes. In this paper, we find that when our data is properly curated we can use \nerfw to recover dense 3D structure. We then show that this structure can be used to guide consistent 3D human reconstruction.

\subsection{Humans in 3D scenes}
Most works that reconstruct humans in context with the scene assume a static, pre-captured 3D environment. Savva~\etal~\cite{savva2016pigraphs} is one of the first works that explore 3D human-scene interactions from RGB-D video, while Hassan~\etal~\cite{hassan2019resolving} study the recovery of 3D humans in context with their environment from monocular images. Many works incorporate environmental constraints for motion estimation from videos~\cite{rempe2021humor,rempe2020contact,shimada2021neural,shimada2020physcap,xie2021physics,yuan2021simpoe,zhang2021learning}, by assuming known floor or contact points. Recently, Guzon~\etal~\cite{guzov2021human} proposed a system for localizing a person in a known environment and estimating their 3D pose. Again, the environment is reconstructed \textit{a priori} and the approach also requires an egocentric sensor and IMUs for pose estimation. Liu~\etal~\cite{liu20204d} propose a method that reconstructs the scene and the people together using egocentric video captured in static outdoor scenes. In this work, we reconstruct structure from much more challenging dynamic scenes, by aggregating 3D information over video content.

Some works~\cite{weng2021holistic,zhang2020perceiving} have studied human reconstruction from single images, while also recovering aspects of the environment. PHOSA~\cite{zhang2020perceiving} recovers humans interacting with objects from in-the-wild images, and is followed by~\cite{weng2021holistic,xu2021d3d} in other settings. While they focus on visible human-object interactions, we consider cases where the scene might not be fully visible. Knowing camera parameters is an integral part of scene perception. SPEC~\cite{kocabas2021spec} regresses camera parameters from a single image. In contrast, we can recover more reliable context for cameras by leveraging the whole collection of images from a TV show environment.
\section{Technical approach}
\label{sec:technical}

\begin{figure*}[t]
  \centering
  \includegraphics[width=\linewidth]{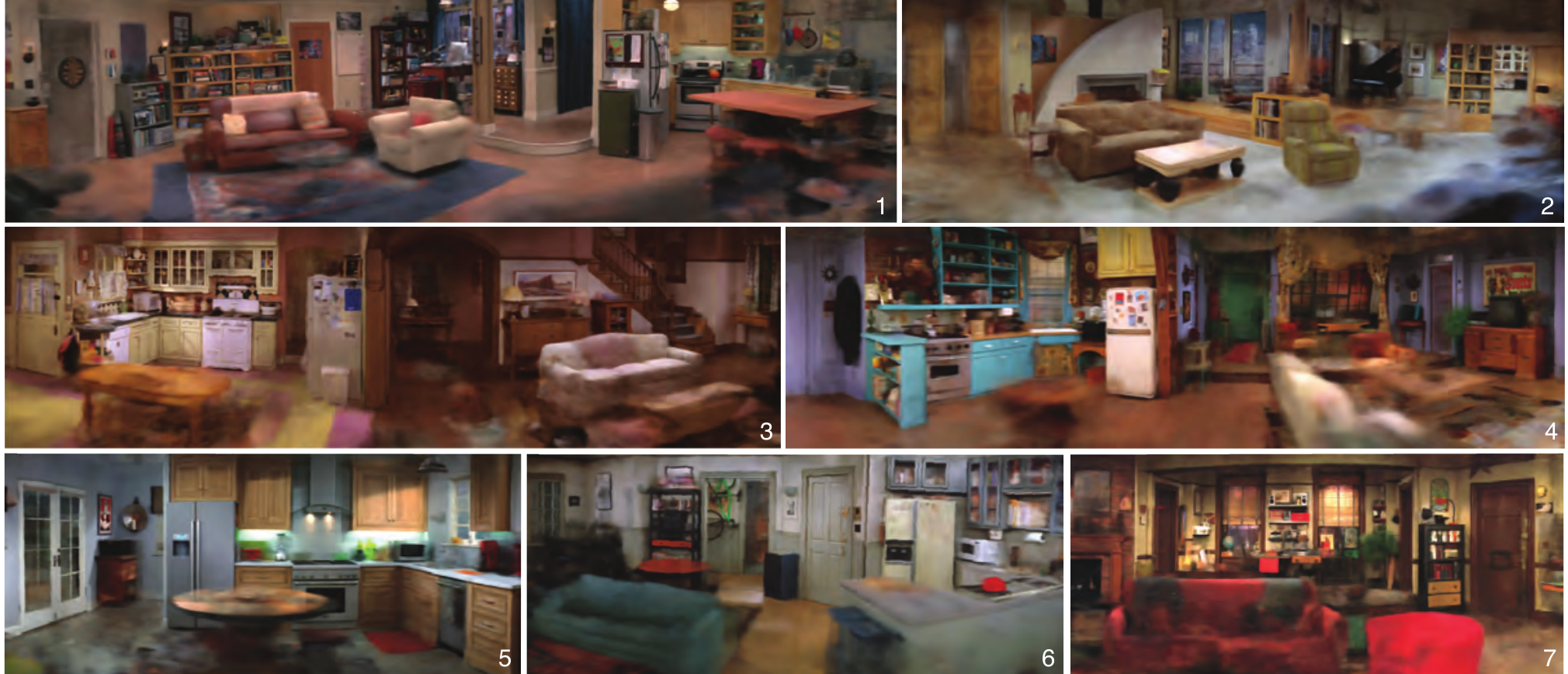}
  \caption{\textbf{Panoramic views of the reconstructed TV show environments.} We obtain and render the static structure using \nerfw~\cite{martin2021nerf}. The environments represent seven TV shows: ``The Big Bang Theory'', ``Frasier'', ``Everybody Loves Raymond'', ``Friends'', ``Two And A Half Men'', ``Seinfeld'' and ``How I Met Your Mother''.}
  \label{fig:nerf_reconstructions}
\end{figure*}

For the following discussion, we use the term {\it environment} to refer to a location, \ie, a room, kitchen, cafe, etc., that appears often in a TV show. Figure~\ref{fig:nerf_reconstructions} visualizes the panoramic view of the environments we reconstruct in this paper. We use the term {\it shot} for an uninterrupted sequence captured by a camera. Shots are organized in {\it scenes}, which are typically captured in the same environment. Multiple scenes comprise an {\it episode} and multiple episodes are organized into a {\it season}. In this work we collect videos across the whole season of a TV show. 

\subsection{Camera estimation}

For the first step of our workflow, we need to register the cameras in a common coordinate frame (\ie, computing intrinsics and extrinsics) for each environment. This amounts to hundreds of thousands of frames across the season. To keep the number of frames at a practical scale for Structure-from-Motion (SfM) pipelines, we sample frames at shot boundaries, which are automatically detected~\cite{huang2020movienet}. This helps to increase the variety of viewpoints - we only use two frames per shot, and inter-shot variety is typically larger than intra-shot variety.

On this reduced set of frames, we use DISK~\cite{tyszkiewicz2020disk} to find correspondences. Since our data includes dynamic actors, we run Mask R-CNN~\cite{he2017mask} to detect human masks, and we reject correspondences on these regions. We use COLMAP~\cite{schonberger2016structure} on the remaining feature matches and estimate the sparse 3D reconstruction and camera registration. We use a simple pinhole camera model, and allow each camera to have different focal length. For each frame $t$ we get estimates of camera intrinsics $K_t \in \mathbb{R}^{3\times3}$ and extrinsics $R^{\CWsmall{}}_t \in \mathbb{R}^{3\times3}, T^{\CWsmall{}}_t \in \mathbb{R}^{3}$, where $CW$ denotes camera to world transformation. This sparse reconstruction is used to register other frames (non shot-boundary images). Since we do not have access to 3D ground truth for TV show environments, the quality of our cameras is evaluated implicitly by the effect it has on the human reconstruction (see also Sup. Mat.).

\subsection{Dense structure}

Besides the camera registration returned from SfM, we also estimate the dense structure of the environment to help with human position estimation. Traditional dense reconstruction methods assume static scenes, but these assumptions are not satisfied in TV show environments which contain many images of extreme diversity (Figure~\ref{fig:in_the_wild}). Instead, we use a \nerfw network~\cite{martin2021nerf} for dense structure estimation. \nerfw extends NeRF~\cite{mildenhall2020nerf}, to account for varying appearances and transient occluders. For efficiency, instead of training \nerfw with all images, we use an automatic selection method to maximize viewpoint variety. We cluster the images based on camera location and viewing direction. For each cluster we select the image with least percent of Mask R-CNN human pixels to use for training (\ie, maximum number of scene rays). After training, \nerfw returns a volumetric 3D representation of the static structure of the scene.

\begin{figure}[t]
\centering
\includegraphics[width=\linewidth]{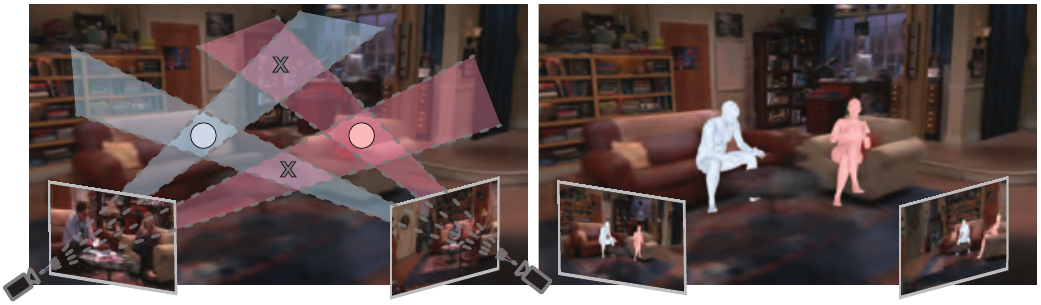}
\caption{\textbf{Calibrated cameras for scale estimation and identity association.} Given calibrated cameras, we can use frames at a shot change to solve for the actors' pose, location, relative scale and association. The four overlapping regions (left) indicate possible locations triangulated by the cameras. Circles indicate correct matches after Hungarian matching.
Reconstructed humans are visualized in a NeRF (right).}
\label{fig:scale_and_position}
\end{figure}

\subsection{Calibrated multi-shot human reconstruction}
\label{sec:cmshot}
In movies and TV shows, scenes are filmed in consecutive shots. The shot changes within a scene correspond to consecutive time frames seen by different viewpoints. This serves as {\it effective multi-view} information, providing signal to recover the 3D location and pose of the actors~\cite{pavlakos2020human}. However, doing so requires knowledge of the identity of the actors across the shot changes. Prior work utilizes a pre-trained recognition-based re-ID model to establish these correspondences, but this is not always reliable, for example when only the back of the character is visible. We make an observation that when camera information is available, the association can be solved jointly with the 3D human pose, shape, and location. We refer to this approach as \textit{calibrated multi-shot optimization}.

Let us assume there are $M$ actors in frame $t$, $N$ actors in frame $t+1$, and a shot change happens from frame $t$ to $t+1$. We need to solve a matching problem to associate the two sets of actors. We propose to use the objective of SMPLify fitting~\cite{bogo2016keep} to model the cost for this matching.

Formally, let us consider a detection of a person at time instance $t$, with detected 2D keypoints $J_{est,t}$~\cite{cao2019openpose}. We denote with $\theta_t$ the pose parameters and with $\beta_t$ the shape parameters of the person in the SMPL format~\cite{loper2015smpl}. We use $J_t$ for the joints and $T^{\Csmall{}}_t$ for the translation of the body in the {\it camera frame}. Moreover, from SfM, we have access to the transformations $R^{\CWsmall{}}_t, T^{\CWsmall{}}_t$ from the camera frame to the world frame at time $t$. Given all of the above, we minimize the objective function with respect to $\{\theta_t, \theta_{t+1}, \beta_t, \beta_{t+1}, T^{\Csmall{}}_t, T^{\Csmall{}}_{t+1}\}$:
\begin{equation}
E = \underbrace{E_{J_t} + E_{J_{t+1}}}_\textrm{2D reprojection} + 
\underbrace{E_{{\textrm{priors}}_{t}} + E_{{\textrm{priors}}_{t+1}}}_\textrm{anthropometric constraints} +
\underbrace{E_{{\textrm{glob}}_{t,t+1}}}_\textrm{3D consistency}
\end{equation}
Here, $E_{J_t} = E_{J_t}(\beta_t,\theta_t,K_t,J_{est,t})$ is the joints reprojection term and $E_{\textrm{priors}}$ are anthropometric priors similar to~\cite{bogo2016keep}. The key constraint is multi-shot consistency, which encourages the estimated bodies to be similar in the global frame:
\begin{equation}
E_{\textrm{glob}_{t,t+1}} = \|(R^{\CWsmall{}}_t J^{\Csmall{}}_t +T^{\CWsmall{}}_t) - (R^{\CWsmall{}}_{t+1} J^{\Csmall{}}_{t+1} + T^{\CWsmall{}}_{t+1})\|^2.
\end{equation}
In contrast to prior work, we do not need to solve for the camera as we have access to reliable extrinsics and intrinsics (prior works~\cite{jiang2020coherent,pavlakos2020human,zhang2020perceiving} use a heuristic for focal lengths). This leads to more accurate human placement and constraints that allow for solving associations. Using this fitting cost $E$, we solve association by Hungarian matching. See Figure~\ref{fig:scale_and_position} for an illustration of this optimization.

\begin{figure*}[t]
\centering
\includegraphics[width=\linewidth]{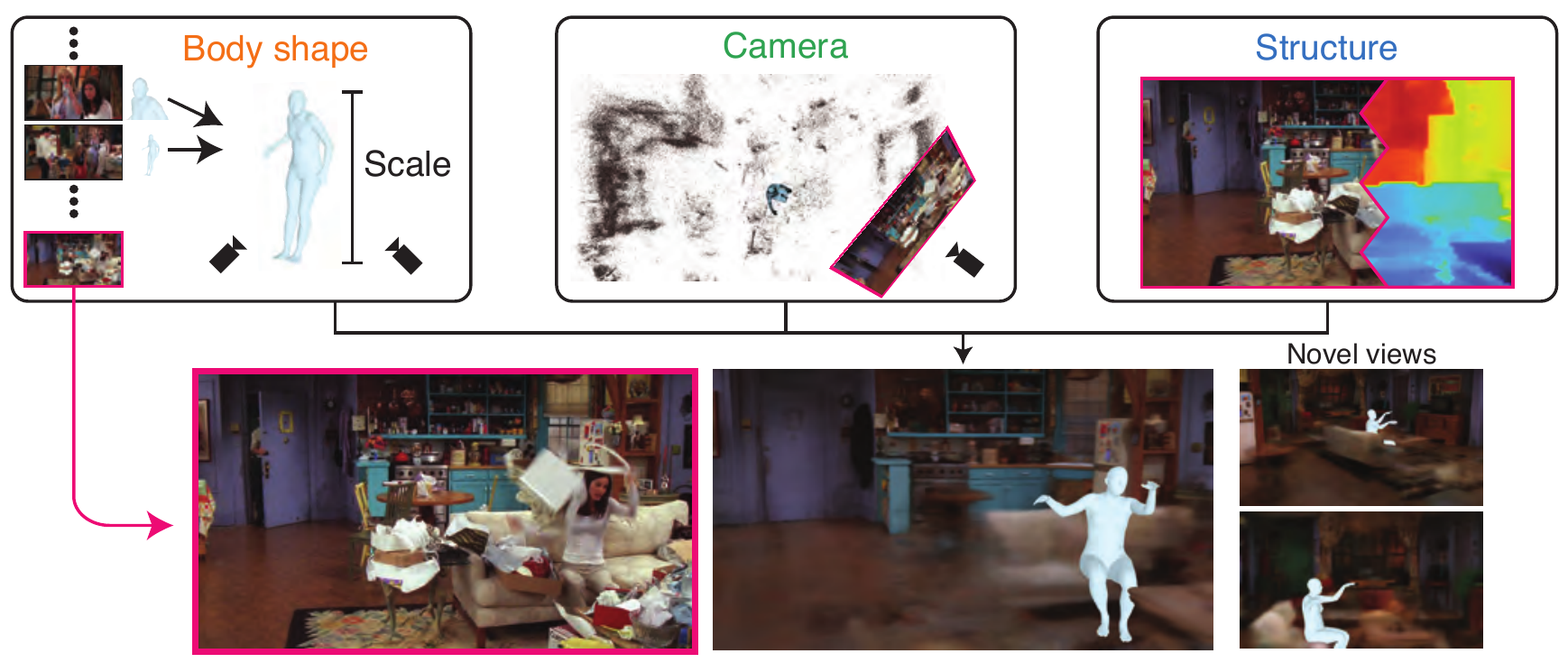}
\caption{\textbf{Contextual monocular human reconstruction.} For an input frame, we can leverage (a) the \scale{body shape} (scale) of the person from a neighboring shot change, (b) the \camera{camera} registration, and (c) the static \structure{structure} of the environment. This enables monocular reconstruction of the person in context with their environment.}
\label{fig:contextual_reconstruction}
\end{figure*}

\subsection{Contextual monocular human reconstruction}

Although shot changes provide effective multi-view information for free, the majority of the frames in the video only have monocular observations. Monocular human reconstruction is challenging, particularly so for TV shows with many close-up shots; however, in our case, we can capitalize on the contextual information we have recovered. In this subsection, we explain how we can make use of this 3D context in an effective way. We demonstrate this using a single-frame optimization approach, SMPLify~\cite{bogo2016keep}, but other methods could also benefit from our context, \eg, we show representative results for HuMoR~\cite{rempe2021humor} in the Sup. Mat.

A high-level overview of this step is presented in Figure~\ref{fig:contextual_reconstruction}. First, given the sparse reconstruction of the environment, we can register the \camera{camera} for a new frame. This gives us both extrinsics \camera{$R^{\CWsmall{}}_t$}, \camera{$T^{\CWsmall{}}_t$} and intrinsics \camera{$K_t$} for the camera via solving PnP with COLMAP~\cite{schonberger2016structure}. We leverage these parameters for accurate projection. Moreover, we can employ the structure captured by our \nerfw network. In general, it is not trivial to extract the structure from NeRF~\cite{oechsle2021unisurf,yariv2021volume}. The native representation used by NeRF is in the form of densities for each point. Here, we propose to use this density as a proxy for occupancy of the 3D space. With this in mind, we formulate an objective to discourage the human body vertices $V$ from occupying areas with high density values:
\begin{equation}
E_{\structure{\textrm{structure}}} = \rho \Bigl( \sum_{v \in V} \structure{$\Tilde{\sigma}$}(v) \Bigr ),
\end{equation}
where $\Tilde{\sigma}$ samples values from the density field $\sigma$ using trilinear interpolation, while $\rho$ is the Geman-McClure robust error function~\cite{geman1987statistical}. Finally, we leverage the shape parameters \scale{$\hat{\beta}$} that capture the relative scale of the person with respect to the environment, and are recovered from the nearest shot change with the calibrated multi-shot reconstruction. This value can be used explicitly in the optimization to resolve the scale ambiguity.

Eventually, our monocular fitting objective minimizes:
\begin{equation}
E_{J}(\beta=\scale{$\hat{\beta}$},\theta,K=\camera{$\hat{K}$},J_{est}) + E_{\textrm{priors}} + E_{\structure{\textrm{structure}}},
\label{eq:data_term}
\end{equation}
with respect to ${\theta_t, T_t^{\Csmall{}}}$, where we employ the \camera{camera} information and the \scale{body shape} parameters of the person during the fitting, while also discouraging the body mesh from penetrating the static \structure{structure} of the scene.

\subsection{Applications}
An important argument in favor of 3D reconstruction for people in TV show environments is that it can simplify many reasoning tasks in this domain. For example, the calibrated multi-shot optimization explicitly reasons about the identity of the detected humans, as part of the Hungarian matching. This enables reliable {\it re-identification} in the challenging case of shot changes where the viewpoint can change significantly (Section~\ref{sec:experiments_reid}). Moreover, one can extract {\it gaze information} from our 3D humans by considering the 3D pose of the face/head. With knowledge of camera pose, we can easily estimate the gaze direction in the global space, and thus compute gaze across the shot change (Section~\ref{sec:experiments_gaze}). Finally, we perform an analysis of our data which could be useful for {\it cinematography} applications, and highlight the potential of {\it image editing} using our results (Section~\ref{sec:analysis}).
\section{Experiments}
\label{sec:experiments}

In this section, we present the quantitative and qualitative evaluation of our approach. We use seven popular TV shows (Figure~\ref{fig:nerf_reconstructions}) and one season from each. We follow the procedure described in Section~\ref{sec:technical} to collect the images we use. Each environment has 1k-5k frames from shot changes. For evaluation, we select per TV show a set of 50 person identities present on these shot changes. We use these frames as a test set to evaluate our method qualitatively with a crowd-sourced perceptual evaluation on AMT and curate it with the information we require for quantitative evaluation, \ie, human-human associations, body keypoints, top-down location of the pelvis in the scene and gaze target across the shot change.

\begin{figure*}[t]
\centering
\includegraphics[width=\linewidth]{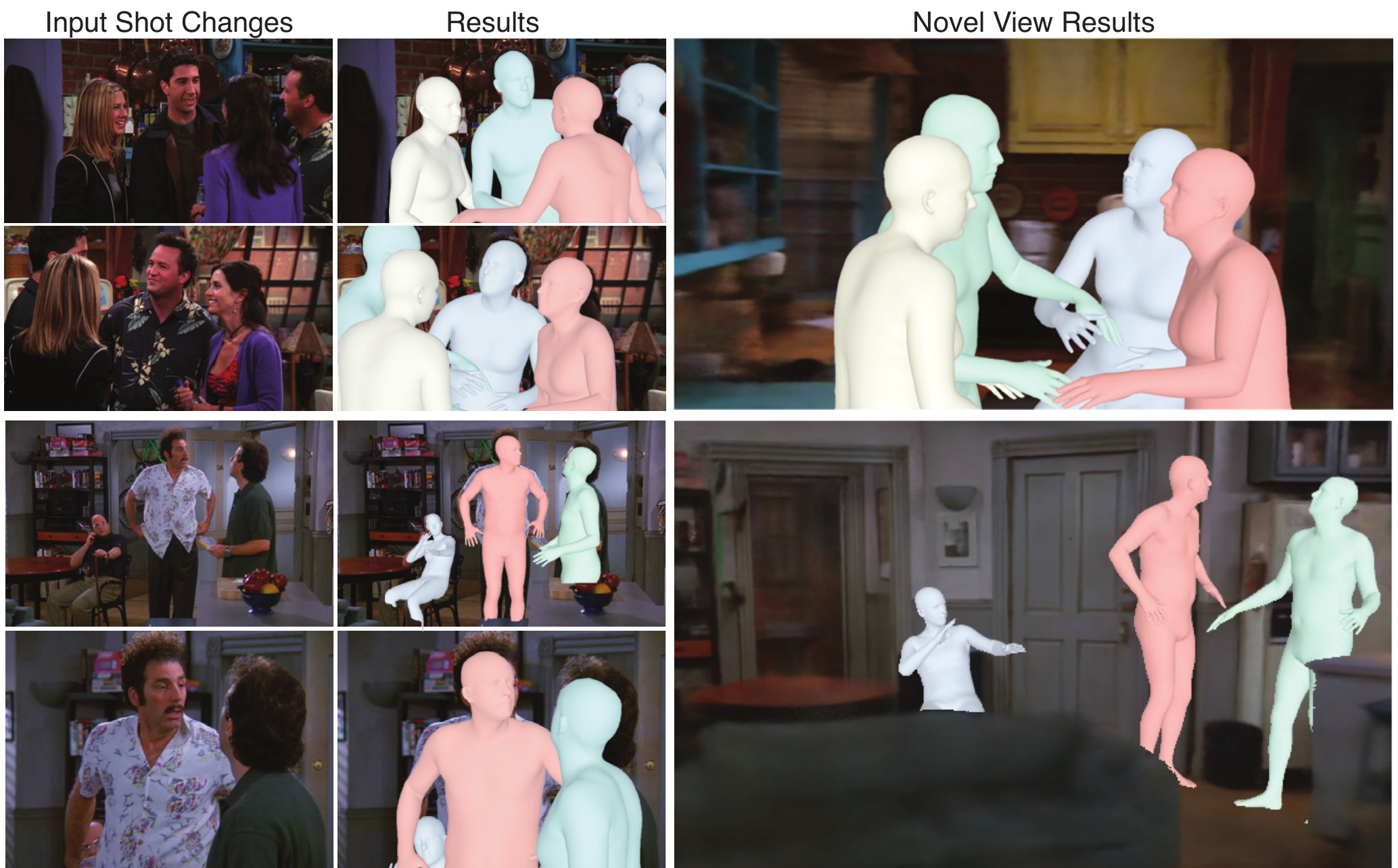}
\caption{\textbf{Calibrated multi-shot and re-ID results.} Using input shot changes (left), we perform our calibrated multi-shot optimization which jointly solves for pose, shape, location and association (middle). Note that identity, illustrated with colors, is not available a priori, but is estimated jointly with the 3D reconstruction. The recovered humans can be rendered in novel views using the NeRF of the environment (right).}
\label{fig:calibrated_multishot_reid}
\end{figure*}

\subsection{Calibrated multi-shot human reconstruction}
For a proof of concept, we first evaluate our proposed calibrated multi-shot optimization in a controlled setting, with the Human3.6M dataset~\cite{ionescu2013human3}, where we have accurate 3D ground truth for pose. Since our focus is on the effect of having access to camera parameters, we compare with the equivalent uncalibrated baseline, which is similar to~\cite{pavlakos2020human}. The results are presented in Table~\ref{tbl:calibrated_mshot}. The significant improvement when having access to camera information further motivates the importance of our calibrated multi-shot algorithm.

For our data from TV shows, we do not have access to 3D ground truth for humans, so we perform two evaluations for the human reconstructions. First, we perform a system evaluation by Amazon Mechanical Turk (AMT) workers. For each 3D human reconstruction, we task the annotators to select the rendered result video (our method vs. a baseline) where the human reconstruction is more accurate and consistent with the scene and shot boundary images. Each result video is 10 seconds and provides multiple viewpoints of the person in the scene. We test on our test set, resulting in 2100 human labels from 48 participants who went through quality control (please see Sup. Mat. for more details). We report the percent of choices where our method is preferred over the baselines in Table~\ref{tbl:calibrated_mshot}. The uncalibrated baseline (first row; without intrinsics or extrinsics) is very rarely preferred over our calibrated baseline (last row; with estimated intrinsics and extrinsics). Having access to estimated intrinsics can help with localization (middle row), but it is still preferred only 35\% of the time. Besides the crowd-sourced evaluation, we also evaluate the location of each person quantitatively. In this case, we compute the mean metric distance error for the pelvis joint in the top-down projection, which is reported in Table~\ref{tbl:calibrated_mshot}. The conclusions are consistent with the AMT evaluation, highlighting the importance of camera information. Some representative results of our calibrated multi-shot optimization are presented in Figure~\ref{fig:calibrated_multishot_reid}, where we also indicate the estimated identity association (which we evaluate in more detail in Section~\ref{sec:experiments_reid}).

\begin{figure*}[t]
\centering
\includegraphics[width=\linewidth]{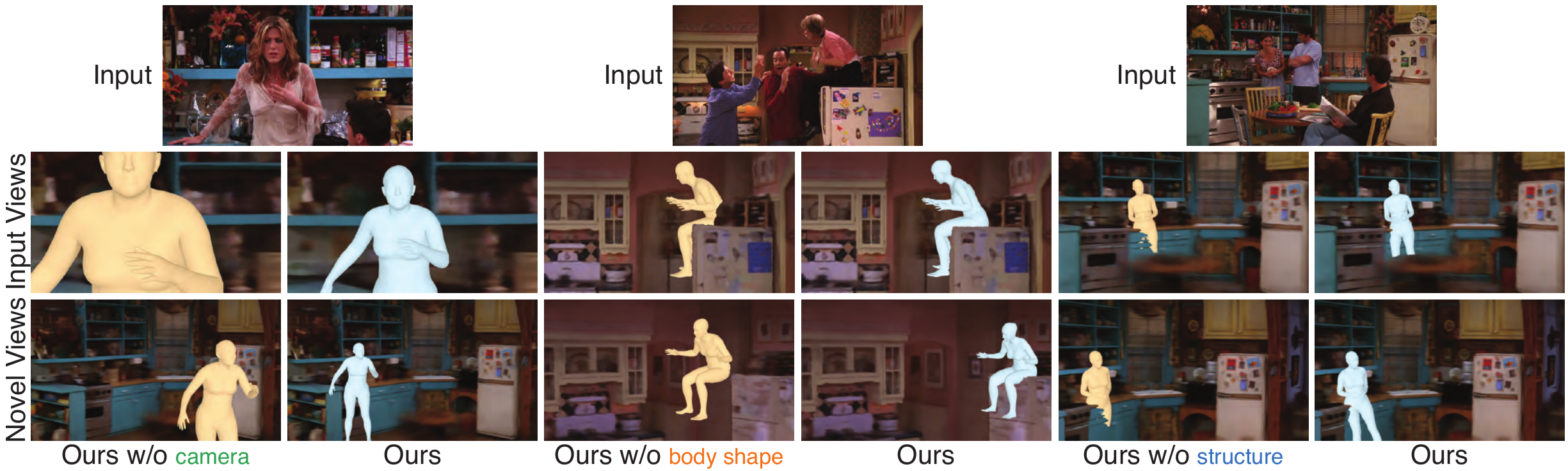}
\caption{\textbf{Results for the contextual monocular reconstruction.} We ablate the basic components of the contextual reconstruction to demonstrate their effects. Our method uses all three forms of context. Without our estimated camera intrinsics (left) and without body shape (middle), the person is incorrectly placed in the scene due to scale ambiguity. Using structure (right) avoids interpenetration with the environment.}
\label{fig:contextual_reconstruction_results}
\end{figure*}

\begin{table}[b]
\centering
\begin{tabular}{l|cc|cc|cc}
\toprule[0.4mm]
Method & \multicolumn{2}{c}{Camera information} & \multicolumn{2}{c}{Human3.6M} & \multicolumn{2}{c}{TV shows}  \\ \hline
Multi-shot & \multirow{2}{*}{Intrinsics} & \multirow{2}{*}{Extrinsics} & \multirow{2}{*}{MPJPE} & \multirow{2}{*}{PA-MPJPE} & \% preferred & Distance \\
optimization & & & & & vs. Ours $\uparrow$ & error $\downarrow$ \\ \hline
Uncalibrated~\cite{pavlakos2020human} & \xmark & \xmark & 131.9 & 56.9 & 4\% & 889cm \\
Partial Calibration & \cmark & \xmark & 123.8 & 56.3 & 35\% & 59cm \\
Calibrated & \cmark & \cmark & {\bf 65.8} & {\bf 47.1} & --- & {\bf 38cm} \\
\bottomrule[0.4mm]
\end{tabular}
\caption{{\bf Evaluation of the proposed calibrated multi-shot optimization.} We ablate the effect of camera information in multi-shot optimization. On Human3.6M, we report results on the standard 3D pose metrics in mm~\cite{zhou2018monocap}. On our TV show data, we perform a system evaluation on AMT and provide quantitative results based on the spatial localization of the reconstructed person in the scene.}
\label{tbl:calibrated_mshot}
\end{table}

\subsection{Monocular contextual human reconstruction}
Next, we investigate our proposed contextual monocular reconstruction. For this evaluation, we study the effect of each component separately -- knowledge of the \camera{cameras}, access to the person's \scale{body shape}, and finally scene \structure{structure} information. We present the results of this ablation in Table~\ref{tbl:contextual_ablation}, where we report cross-shot PCK @ $\alpha=0.5$~\cite{pavlakos2020human}. Effectively, we project the person to the view across the shot boundary and measure localization accuracy for the joints in that space (more details in the Sup. Mat.). First, we see that state-of-the-art monocular methods without context~\cite{bogo2016keep,kocabas2021pare,kolotouros2021probabilistic} perform similarly on this data. Then, we examine the effect of context, using the optimization baseline~\cite{bogo2016keep} as our starting point (third row). Access to camera intrinsics is important to estimate a rough location of the person, and without it the method performs as the baseline without context. Knowledge of the body scale of the person, can make our estimate even more accurate. Finally, structure gives a smaller quantitative improvement but has a more pronounced qualitative effect by placing the person coherently in the environment. See Figure~\ref{fig:contextual_reconstruction_results} for qualitative results.

\begin{table}[t]
\centering
\begin{tabular}{l c}
\toprule[0.4mm]
Method & cross-shot PCK \\ \midrule
No context: ProHMR~\cite{kolotouros2021probabilistic}  & 14.7\% \\
No context: PARE~\cite{kocabas2021pare}    & 14.2\% \\ \hline
No context: SMPLify~\cite{bogo2016keep}    & 16.5\% \\ \hline
Context w/o \camera{camera} (intrinsics)   & 16.0\% \\
Context w/o \scale{body shape} (scale)     & 65.9\% \\
Context w/o \structure{structure}          & 87.5\% \\ \hline
Context (full)    & {\bf 88.7\%} \\
\bottomrule[0.4mm]
\end{tabular}
\caption{{\bf Ablation of the main components of our contextual reconstruction.} Cross-shot PCK @ $\alpha=0.5$ is reported. Knowledge of the camera focal length is very important to get a good 3D location for the human. Information about body shape can have significant improvements, as it resolves the scale ambiguity. Structure helps to avoid the incoherent interpenetrations with the scene.}
\label{tbl:contextual_ablation}
\end{table}

\begin{table}[b]
\centering
\begin{tabular}{lc}
\toprule[0.4mm]
Matching costs & Re-ID F1 $\uparrow$ \\ \midrule
Fu~\etal~\cite{fu2021unsupervised} (Appearance) & 0.78 \\
Huang~\etal~\cite{huang2018person} (Appearance) & 0.79 \\
Huang~\etal~\cite{huang2020movienet} (Appearance) & 0.80 \\
Keypoint triangulation (Geometry) & 0.86 \\
Ours (Geometry + Anthropometric) & \textbf{0.91} \\
\bottomrule[0.4mm]
\end{tabular}
\caption{{\bf Re-ID results for actors in shot boundary frames.}
We use different methods to estimate matching costs for detections and we run Hungarian matching to establish associations. A geometric baseline using the reprojection error from person keypoint triangulation improves upon SOTA image-based baselines~\cite{fu2021unsupervised,huang2018person,huang2020movienet}, but using our multi-shot fitting cost performs better because it also includes anthropometric constraints, \ie, the triangulated points should respect the human body priors.}
\label{tbl:reid}
\end{table}

\subsection{Re-identification}
\label{sec:experiments_reid}

For the re-ID evaluation, we examine the challenging case of person association after a shot change. For our case, re-ID is directly estimated from our calibrated multi-shot optimization. We compare this result with two types of baselines for computing affinities/costs between instances for Hungarian matching. The first type is image-based re-ID networks for affinity estimation, where \cite{fu2021unsupervised} achieves SOTA on standard re-ID benchmarks, while \cite{huang2018person,huang2020movienet} are trained on movies, a source of data similar to TV shows. The second type is a geometric baseline that uses our recovered cameras and is based on human keypoint triangulation, where the reprojection error is used as the cost for the Hungarian algorithm. Notice, that unlike SMPL fitting, this does not incorporate anthropometric constraints, \ie, it considers every keypoint match independently, without using human body shape priors or measuring the holistic result. We report re-ID F1 scores in Table~\ref{tbl:reid} using the visible pairs of actors before/after the shot change. Based on the results, our re-identification can consistently outperform these baselines.

\begin{wraptable}{R}{0.5\textwidth}
\centering
\begin{tabular}{lcc}
\toprule[0.4mm]
\multirow{2}{*}{Method} & \multicolumn{2}{c}{PCGD ($\alpha=20^o$) $\uparrow$} \\
 & all & w/ face \\ \midrule
Recasens~\etal~\cite{recasens2017following} & 16\% & 32\%  \\
Ours & {\bf 62\%} & {\bf 67\%} \\
\bottomrule[0.4mm]
\end{tabular}
\caption{{\bf Gaze following results.} We report the Percentage of Correct Gaze Directions (see text for description). Our approach outperforms the baseline of~\cite{recasens2017following}.}
\label{tbl:gaze}
\end{wraptable} 

\subsection{Gaze estimation}
\label{sec:experiments_gaze}
For gaze estimation, we compare with the method of~\cite{recasens2017following} that estimates the gaze target after the shot change. We evaluate the angular error in the gaze direction projected on the image plane. We report the Percentage of Correct Gaze Directions (similar to PCK~\cite{yang2012articulated}), using $\alpha=20^o$ as threshold. Please see Sup. Mat. for details.

Results are reported in Table~\ref{tbl:gaze}. Since~\cite{recasens2017following} relies on face detection, we report results on our whole test set (column ``all'') and on the subset where face detection is successful (column ``w/ face''). Our approach outperforms~\cite{recasens2017following} in both cases. Since we rely on body detection, we are more robust even when the face is occluded. Moreover, our extracted camera poses allow us to follow gaze across shots in a more accurate way. Further improvements are expected by modeling eye pose and saliency estimation to detect the gaze target, similarly to~\cite{recasens2017following}

\subsection{Cinematography/Image editing applications}
\label{sec:analysis}

\begin{figure}[t]
\centering
\includegraphics[width=\linewidth]{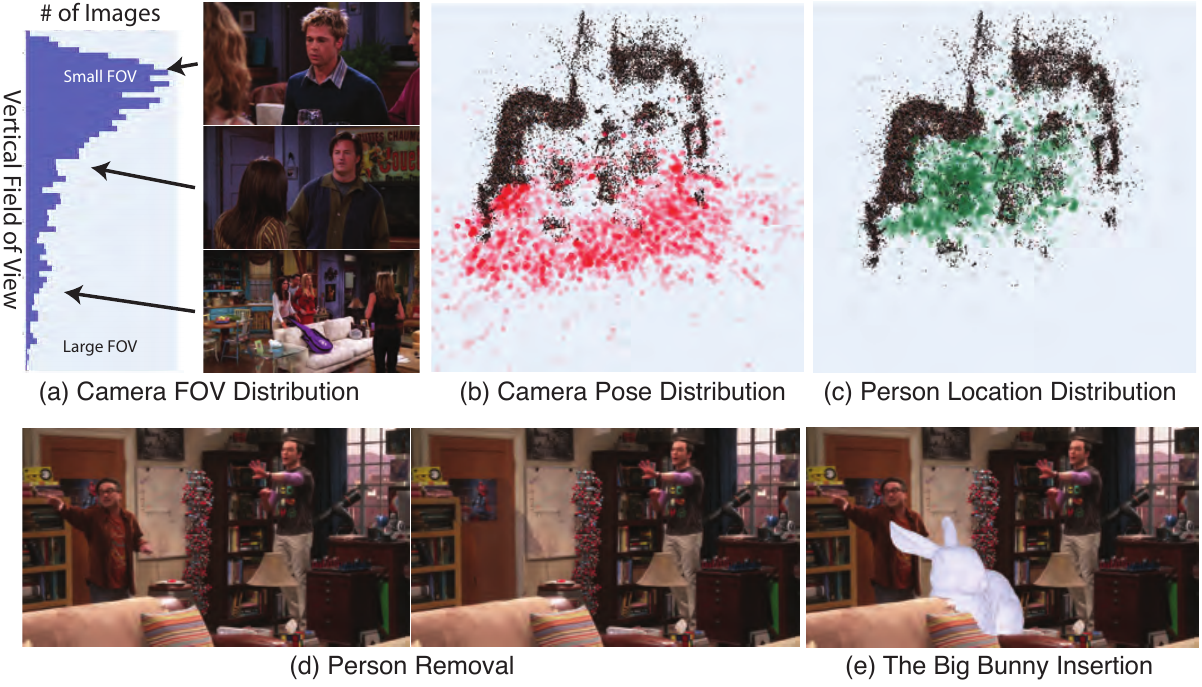}
\caption{\textbf{Cinematography applications/Image editing.} We present analysis of our processed data, including distribution of field of view, camera pose distribution and person location distribution for Friends (top). Moreover, we present editing options after our processing, including person removal and object insertion (bottom).}
\label{fig:analysis}
\end{figure}

We provide an initial analysis of our results in Figure~\ref{fig:analysis}, and present an extended study in the Sup. Mat. First, we visualize the \textit{distribution of the estimated field of view for the cameras}. Here we can see the long-tail distribution for the views with a large field of view, \ie, more informative viewpoints for 3D reconstruction. This justifies our insight to process data across the whole season, since the large majority of the views is typically close-ups. Moreover, we visualize the \textit{locations of the cameras and the human actors}. The camera data could be useful for cinematography analysis~\cite{savardi2018shot}, and the person data for behavior or affordance analysis~\cite{wang2017binge}. Finally, we illustrate potential editing applications enabled by the reconstruction of humans and the environment: \textit{person removal and object insertion}. More editing options are possible, given the 3D nature of our processing.

\section{Discussion}
\label{sec:conclusion}

\noindent
{\bf Conclusion}:
To the best of our knowledge, we are the first to reconstruct the people and the environment in TV shows and reason about them in 3D. We start with multi-shot video sequences associated with a specific environment and recover the camera, structure and relative human scale. We use this information as context to reconstruct humans even from a single frame, in a way that is consistent with their environment. We demonstrate our approach on seven different TV shows and present qualitative and quantitative results, as well as a wide variety of applications and analysis of the reconstructed data.

\noindent
{\bf Future Directions}: Our work has only scratched the surface of this extremely challenging and in-depth problem. Currently, we do not reconstruct the transient objects or dynamic objects that humans interact with (\eg, chairs that move around, fridge opening). Also, the recovered pose of the humans is completely dependent on the quality of the 2D keypoint detections. It would be an interesting direction to incorporate appearance models for pose fitting. 

\noindent
{\bf Acknowledgements:} This research was supported by the DARPA Machine Common Sense program as well as BAIR/BDD sponsors. Matthew Tancik is supported by the NSF GRFP.

\clearpage
%
%
\bibliographystyle{splncs04}
\bibliography{egbib}
\end{document}